\documentclass[10pt, a4paper]{article}

\usepackage[final]{lrec2026} 

\usepackage{url}            
\usepackage{booktabs}       
\usepackage{amsfonts}       
\usepackage{nicefrac}       
\usepackage{microtype}      
\usepackage{xcolor}         
\usepackage{amsmath}
\usepackage{multirow}
\usepackage{arydshln}
\usepackage{amssymb}
\usepackage{cleveref}
\usepackage{comment}
\usepackage{colortbl}

\usepackage{subfig}

\title{TrackList: Tracing Back Query Linguistic Diversity for Head and Tail Knowledge in Open Large Language Models}

\name{Ioana Buhnila\textsuperscript{1} \textsuperscript{2}, Aman Sinha\textsuperscript{1} and Mathieu Constant\textsuperscript{1}} 

\address{\textsuperscript{1}ATILF, University of Lorraine - CNRS, France\\
         \textsuperscript{2}Center for Data Science in Humanities, Chosun University, South Korea\\
         firstname.lastname@univ-lorraine.fr\\}

\abstract{
Large Language Models (LLMs) have proven efficient in giving definition-type answers to user input queries. While for humans giving various types of answers, such as examples and paraphrases, is an easy task, LLMs struggle to provide correct answers for other than definition-type queries. In this study, we evaluated this drop in performance using \texttt{TrackList}, a fine-grained linguistic and statistical analysis pipeline to investigate the impact of the pre-training data on LLMs answers to diverse linguistic queries. We also introduce \texttt{RefoMed-EN}, an English dataset consisting of 6170 human-annotated medical terms alongside their corresponding definitions, denominations, exemplifications, explanations, or paraphrases. We studied whether the high frequency of a concept (head) or low frequency (tail) impacts the language model's performance. We evaluated the quality of the LLM's output using syntactic and semantic similarity metrics, statistical correlations and embeddings. Results showed that the LLM's task performance for definition type questions is the highest, while for the exemplification type it is the lowest. Additionally, we showed that for definition-type questions, large language models are prone to paraphrase more on popular and frequent knowledge and less on tail and technical knowledge, especially in the expert texts.
 \\ \newline  }

\begin{document}

\maketitleabstract

\section{Introduction}
As Large Language Models are becoming widely used in many writing tasks and are applied to different fields, the need to explain and interpret the generated responses has become crucial. In order to be trusted and used in sensitive fields such as medicine, finance or law, understanding the real capabilities of LLMs is indispensable. Methods such as In-Context Learning (ICL) \cite{dong2024survey}, Chain-of-Thought (CoT) prompting \cite{wu2023analyzing}, and Retrieval Augmented Generation systems (RAG) \cite{lewis2020retrieval} have contributed to further interpret the LLM's generated response, and help mitigate hallucinations \cite{akbar2024hallumeasure,huang2024survey, asai2023self}. Black box models like ChatGPT and other proprietary models showed limitations in terms of explainability and reproducibility \cite{zhao2024explainability, liesenfeld2023opening, ravichandran2024xai}.

\begin{figure}[t]
    \centering
    \includegraphics[trim = 1cm 0.6cm 1cm 0.3cm, scale=0.6]{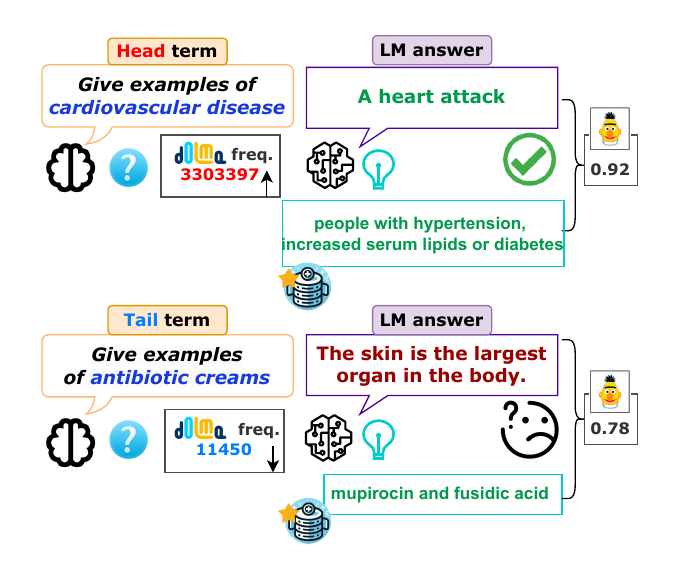}
    \caption{Large Language Models  tend to generate more hallucinated \textit{definition-style} outputs even when directly asked to answer an \textit{example-style} query for tail knowledge terms.}
    \label{qatype}
\end{figure}


\textcolor{black}{In this work we explored non-proprietary and open Large Language Models, like OLMo \cite{groeneveld2024olmo} and Pythia \cite{biderman2023pythia}. We chose these language models (LMs) because they are fully open, thus fostering interpretability and reproducibility. These LLMs were released together with model weights, inference code, training/evaluation code, and pretraining corpus, DOLMA \cite{soldaini2024dolma} for OLMo, and The Pile \cite{gao2020pile} for Pythia. We further investigated OLMo and Pythia's linguistic capabilities by conducting a fine-grained linguistic and statistic analysis of how a language model uses pretraining data in the composition of the generated text.} We tested these models in the most widespread real world use-case: open question-answering, meaning the model gives a free text answer to a user question \cite{shailendra2024survey}. LMs are most efficient in question-answering tasks that cover popular information (head knowledge) usually from the general domain \cite{mallen2023not,li2024role} and when giving definitions to concepts. However, their performance drops when tested on torso or tail knowledge \cite{sun2023head, kandpal2023large} or when tackling complex questions \cite{daull2023complex}.  Our study analyzed the impact of query linguistic diversity on QA performance, and demonstrated that LLMs perform worse when asked to give different types of answers, such as examples, to tail medical knowledge (as shown in Figure 1).

We further explored this research question by applying our experiments to the medical domain, as it contains a wide range of head and tail knowledge concepts, such as \textit{cancer}, \textit{asthma} on the head side, and complex and rare medical concepts such as \textit{ideatory apraxia} or \textit{erythematous angina} for rare medical knowledge. Furthermore, we evaluated the LLM's performance for different types of questions  by exploring term requency in the pre-training data.

\textcolor{black}{Our research questions are the following: \textbf{(RQ1)} \textit{Does the frequency of a concept in the pretraining data influence a language model's answer quality?} \textbf{(RQ2)} \textit{How does the linguistic diversity and complexity of questions impact the language model ability to give accurate answers?} \textbf{(RQ3)} \textit{To what extent are language models paraphrasing the data from their parametric memory for head and tail knowledge?} We hypothesize that frequent (head) terms may lead to distributional memorization, while less frequent terms (tail) may result in lower downstream task performance due to reliance on parametric knowledge \cite{wang2024generalization}.}


To explore these hypothesis, we propose a fine-grained analysis and evaluation pipeline, \texttt{\textbf{TrackList}}, to \texttt{\textbf{tra}}ce ba\texttt{\textbf{ck}} query \texttt{\textbf{li}}nguistic diver\texttt{\textbf{s}}ity of head and \texttt{\textbf{t}}ail knowledge in Open Large Language Models. Our contributions are the following:
\begin{enumerate}
    \item We built and share our fine-grained linguistic analysis pipeline, \texttt{TrackList}, to evaluate how LLMs answer to five different types of questions according to pragmatic functions of human communication (definitions, exemplifications, explanations, denominations and paraphrases) using a non-contaminated benchmark, \texttt{RefoMed-EN}. We traced the LLM's answers back to their pretraining corpora, DOLMA and The Pile, and showed that LLMs paraphrase more on head terms and are less semantically diverse when dealing with tail concepts. We will share the \texttt{TrackList} code with the research community to ensure reproducibility and to foster new analysis and metrics on LLMs' linguistic abilities in diverse open QA settings.
    \item We share the plug-and-play medical QA dataset, \texttt{RefoMed-EN}, an English dataset of 6170 human annotated medical terms alongside with their context, discourse markers, corresponding paraphrases, lexical and pragmatic functions. The dataset will be available with open license.
    \item We conducted a detailed analysis of the LLMs' performance according to the linguistic complexity of the user question. We showed that LLMs have limited understanding of linguistic differences in queries, as they are more prone to giving definition-style answers to all queries, even when explicitly asked to give examples or a paraphrase for a concept.
\end{enumerate}

\section{Related Work}

In the medical domain, LLM's output must be correct, therefore methods that combine ICL and RAG showed promising results, such as improving the accuracy of GPT4-Turbo's answers to oncology questions from 62.5\ to 83.3\% (with ICL) and 79.2\% of questions (with RAG) \cite{iivanainen2024investigating}. Moreover, QA frameworks like Self-BioRAG that answer biomedical questions by reflecting and retrieving relevant documents showed a 7.2\% improvement on average over the state-of-the-art open Small Language Models (SLM) of 7b or less \cite{jeong2024improving}. 

\begin{figure*}[t]
  \centering
  \includegraphics[trim = 0.7cm 0.2cm 0cm 1.3cm, scale=0.9]{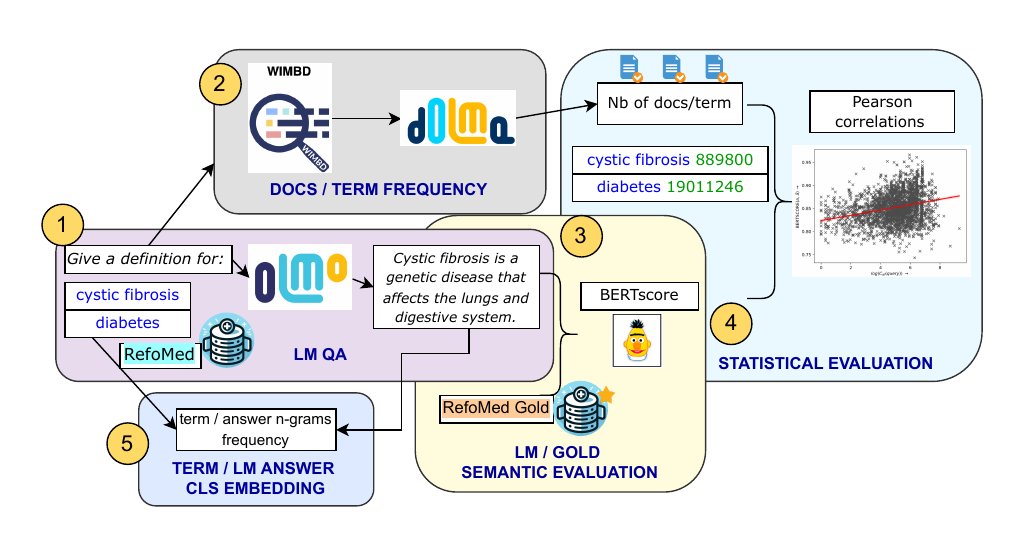}
  \caption{The pipeline of our method represented in five steps. 1) The zero-shot inference QA task using the medical concepts from the RefoMed-EN dataset. The dataset was divided in subdatasets according to the query type (detailed presentation in section 3.4). 2) We obtained the frequency in terms of number of documents for each RefoMed-EN concept. 3) We calculated the BERTScore between the LMs output and the RefoMed-EN gold standard. 4) These two values were used to compute Pearson correlations. 5) We computer a probability metric between the CLS embedding score between the term and the n-grams of the output, and their frequency in the pre-training corpora.}
  \label{fig:method3}
\end{figure*}

Large medical QA datasets like MedQuAD \cite{ben2019question} and MEDIQA \cite{abacha2019overview} gather different types of medical data, such as treatment, symptoms, definition, susceptibility or prevention. These datasets are built for medical experts, more than for the general public interested in understanding medical concepts \cite{nguyen2023medredqa}. Moreover, these datasets are shared in a XML format, not allowing easy implementation for LLMs prompts. Furthermore, the dataset is divided in more medical than linguistic criteria. 

While evaluating LLMs answers to open questions is difficult, recent studies showed that human annotation is still needed when the generated answer does not match a gold standard \cite{kamalloo2023evaluating}. In our work, we combine both human evaluation and automatic evaluation through similarity metrics that we further develop in section 3.2. Determining what is head and tail knowledge is a research question in itself, as recent studies investigated this in detail \cite{mallen2023not,li2024search}. We considered that the word frequency in the pretraining data as the popularity metric for our medical dataset.

Recent studies investigated whether LLMs memorize or generalize knowledge by exploring open language models.  \citeauthor{wang2024generalization} (\citeyear{wang2024generalization}) introduce the concept of distributional memorization, to measure the correlation between the LLM output probabilities and the frequency of the pretraining data. The authors evaluated the task performance with 3 to 5 n-grams search into The Pile \cite{gao2020pile}, the pretraining corpus of the language model Pythia \cite{biderman2023pythia}. They found that LLMs generalize more in reasoning-intensive tasks, while they memorize more in simpler knowledge-intensive tasks \cite{wang2024generalization}. This finding is interesting for our study, as we evaluate how LLMs answer to medical knowledge questions. In this sense, we hypothesize that more frequent knowledge (head) is easier to be memorized by the language model. Concurrent work investigated the concept of linguistic diversity in LLMs answers by evaluating their  lexical, syntactic and semantic distribution compared to human linguistic richness \cite{guo2024benchmarking}.

In our study, we investigated the linguistic complexity of LLM queries from a syntactic, semantic and pragmatic perspective, applied to a knowledge intensive task, medical QA. We present our full method below.

\section{Methodology}
We illustrate our method in Figure~\ref{fig:method3}. \textcolor{black}{We conducted our experiments using four fully open language models, OLMo-1b, OLMo-7b, OLMo-7b-instruct, and Pythia-1b. To access the pre-training corpora, We used the WIMBD tool's API \cite{elazar2023s} for the DOLMA corpus (OLMo) and the \texttt{infini-gram} library \cite{liu2024infini} for The Pile dataset (Pythia).} We counted the number of documents that contain a certain medical concept, thus determining its frequency. We present our detailed rationale in the section below.

\subsection{Formulation} We utilize a language model $\mathbb{M}_\mathrm{LLM}$ and assume access to a corpora $C$ with $\mathcal{D}$ documents which was used to train $\mathbb{M}_\mathrm{LLM}$. We represent the corpora $C$ by a set of unique words $\{w_1, w_2, w_3, ..... w_\infty \}$ present in at least one document $d \in \mathcal{D}$. Each of the document $d \in \mathcal{D}$ can be similarly represented by a subset of $C$ as $\{dw_1, dw_2, dw_3, ..... dw_\infty \}$ where $dw_i$ represents word $w_i$ present in document $d$. We define a document-frequency count operator over corpora $C$ and denote it by $\mathbf{C}_{df}$. It enables us to count the number of documents present in corpora $C$ which contains the provided term at least once\footnote{For example,  $\mathbf{C}_{df}$("\textit{infection}" | $C$) gives 543435456, which implies that there are 543435456 documents in $C$ containing the term \textit{infection} at least once.}.

We further utilize query-answer ($Q,A$) pairs where a question denoted by $q \in Q$ can be of different types as mentioned in \cref{tab:refomed-question-types} and an reference answer $a \in A$ which is borrowed from the RefoMed-EN dataset. In our experiment setup, we provide as input $q \in Q$ to the language model as follows
\begin{equation}
\hat{a} = \mathbb{M}_\mathrm{LLM}(q)
\end{equation}
In the above equation, we denote the generated response from the $\mathbb{M}_\mathrm{LLM}$  as $\hat{a}$. 
During evaluation, we consider different evaluation metrics (denoted by $\Psi$) to calculate the correctness (denoted by $E$) of the generated response $\hat{a}$ with respect to the gold reference $a \in A$ as follows 
\begin{equation}
E = \Psi(\hat{a}, a)
\end{equation}
Lower value of $E$, for example, in case of BERTscore, implies less relevance of generated answer and vice versa.

\begin{table*}
    \centering
    \resizebox{\textwidth}{!}{
    \begin{tabular}{p{3cm}|p{5cm}|p{3.5cm}|p{5.5cm}}
        \toprule
        \rowcolor{blue!15}
       \textbf{Pragmatic function}  &
       \textbf{Definition} &
       \textbf{Query type} & \textbf{Example} \\
         \midrule
        \texttt{Definition}  (\textbf{DEF})& 
        a difficult or technical
        concept is defined to ease comprehension &
        Give a definition for / What is \textit{multiple sclerosis} &
        Multiple sclerosis (MS) is a \textit{disease of the nervous system of immune origin}\\
        \hline
        \texttt{Exemplification} (\textbf{EX}) &
         the meaning of a concept is illustrated through examples of types and subtypes of entities &
         Give examples of \textit{heart and vascular diseases} &
         Heart and vascular diseases include \textit{heart attacks, angina, stroke, sudden cardiovascular death and the need for heart surgery}\\
         \hline
        \texttt{Denomination} (\textbf{DEN})& 
        a concept is reformulated through a semantically similar concept, without simplification &
        Give another denomination for \textit{motor neuron disease} & 
        Pharmacotherapy for pain management in \textit{amyotrophic lateral sclerosis} (motor neuron disease)\\
        \hline
        \texttt{Paraphrase} (\textbf{PARA}) &
        the concept is reformulated through a easy to understand semantically similar synonym &
        Give a paraphrase for \textit{hypotension} &
        Hypotension, i.e. \textit{low blood pressure}, frequently occurs in newborns\\
        \hline
        \texttt{Explanation} (\textbf{EXP}) & 
        the concept is explained through its process or a part of it &
        Give an explanation for
        \textit{autoimmune diseases} &
        These are autoimmune diseases that can be explained by the fact that \textit{the organism produces an antibody against the person’s skin}\\
         \bottomrule
    \end{tabular}}
    \caption{Query types, definitions and examples according to pragmatic functions annotated on the RefoMed-EN corpus. These queries were used for prompting in our QA task.}
    \label{tab:refomed-question-types}
\end{table*}

\subsection{Evaluation Metrics}

\textcolor{black}{As the goal of our study is to advance the interpretability of open large language models, we chose simple and easy to interpret metrics that allow us to analyze the relationship between query term frequency, gold answers, and generated LLM answers.}

\paragraph{Tracing back semantic similarity} We calculated the BERTscore \cite{zhang2019bertscore} between the generated answers and the gold standard from the RefoMed-EN dataset.

\paragraph{Statistical correlation} We computed the Pearson correlation metrics to compare the BERTscore calculated before with the term's frequency in the pretraining corpus.

\paragraph{CLS cosine similarity} We computed the cosine similarity of sentence embeddings to compare the semantic similarity between the medical term and the n-grams of the generated answer.

\subsection{Linguistically Annotated Dataset}


We constructed RefoMed-EN by automatically translating RefoMed\footnote{RefoMed is a French annotated dataset of 6170 annotated medical terms with their corresponding reformulations. The dataset is comprised of paraphrases and reformulations semi-automatically extracted from scientific and popularization medical texts and abstracts from the ClassYN \cite{todirascu2012french} and CLEAR \cite{grabar2018clear} French medical corpora.} \cite{buhnila2024retrieve} from its original version in French to English using a licensed DeepL Translator API \footnote{The dataset required extensive pre-processing as there were formatting and translation inaccuracies from the original annotation.}. We chose to translate this dataset because there is no benchmark annotated on question types following pragmatic and linguistic theories (to the best of our knowledge). The translation of a French dataset assures that there is no benchmark contamination between the pretraining corpora and the test dataset \cite{sainz2023nlp,li2024open}. We will share this new non-contaminated benchmark with the NLP community on github. We present below the question types explored in this study. 

\subsection{Query Linguistic Diversity}

Linguistically, the concept of \textit{reformulation} is defined as textual and discursive act performed with a precise objective \cite{grabar2016we}. \textit{Reformulations} have a well defined pragmatic role by expressing a content in a different semantic or lexical representation, adapted to a specific audience and communicational need, like science popularization or education. The linguistic diversity is correlated to the pragmatic usage of language in humans, such as asking for a definition, an explanation, reformulation or paraphrase of a concept, or to receive examples of a certain entity \cite{grabar2016we}.

In this work, we analyzed the role of reformulations in the case of medical knowledge popularization for laypeople and patients \citeauthor{grabar2016we} (\citeyear{grabar2016we}). We focused on a knowledge intensive question-answering task taking into account the five most common types of questions that require reformulation processes, as shown with examples in Table \ref{tab:refomed-question-types}. We present the number of questions by type from the RefoMed-EN dataset in Table \ref{tab:refomed-question-nb}. 



\begin{table}
    \centering
    \begin{tabular}{rccccc}
        \toprule
        \rowcolor{blue!15}
       &\small{\textbf{DEF}}& \small{\textbf{EX}}&\small{\textbf{DEN}}&\small{\textbf{PARA}}&\small{\textbf{EXP}}\\
         \midrule
        \small{\texttt{ClassYN EX}}&\small{207}&\small{305}&\small{131} & \small{30}&\small{32}\\
        \small{\texttt{CLEAR EX}}&\small{919}&\small{379}& \small{401}&\small{73}&\small{67}\\
        \midrule
        \rowcolor{blue!15}
        \small{\textbf{RefM-EN-EX}}&\small{\textit{1126}}&\small{\textit{684}}&\small{\textit{532}}&\small{\textit{103}}&\small{\textit{99}}\\
        \midrule
        \small{\texttt{ClassYN GP}}&\small{862}&\small{343}& \small{235}&\small{285}&\small{149}\\
        \small{\texttt{CLEAR GP}}&\small{883}&\small{470}&\small{175}&\small{124}&\small{100}\\
        \midrule
        \rowcolor{blue!15}\small{\textbf{RefM-EN-GP}}&\small{\textit{1745}}&\small{\textit{813}}&\small{\textit{410}}&\small{\textit{409}}&\small{\textit{249}}\\
        \midrule
        \small{\textbf{RefM-EN}}&\small{\textbf{2871}}&\small{\textbf{1497}}&\small{\textbf{942}}&\small{\textbf{512}}&\small{\textbf{348}}\\
         \bottomrule
    \end{tabular}
    \caption{Distribution of question types across subcorpora in RefoMed-EN (RefM-EN). EX denotes expert-oriented texts; GP targets the general public. Question types are ordered by frequency (left to right).}
    \label{tab:refomed-question-nb}
\end{table}

\section{Experimental Setup}

This section presents the experiments we conducted to evaluate the generated text according to query types and term frequency. Experiments were done on a P100 NVIDIA GPU, for an individual runtime of $\leq$ 15 hours including different steps involved.

\subsection{Linguistic Diversity and Frequency for Task Performance} We explored the impact the query linguistic diversity has on the quality of the generated answer. We evaluated the task performance of OLMo and Pythia models in a zero-shot QA setting by conducting several experiments:

\begin{itemize}
    \vspace{-0.2cm}\item We prompted the LLM to give a short answer (\texttt{\textbf{a}}) to the given question (\texttt{\textbf{q}}) as an expert in the field: \texttt{"You are a medical expert. Answer the following question in a short sentence"}. The queries were divided by question type, as shown in Table \ref{tab:refomed-question-nb}.
    \vspace{-0.2cm}\item We evaluated the quality of the generated answers for each type of query by computing the BERTscore between the generated answer and the gold standard, the human annotated paraphrases and definitions from RefoMed-EN.
    \vspace{-0.2cm}\item We calculated the Pearson correlation between the BERTscore and the frequency of the medical concept in the pretraining corpora.

\end{itemize}

\begin{table}[ht!]
    \centering
    \begin{tabular}{c ccccc}
    \toprule
       & \textbf{DEF}  & \textbf{EX}& \textbf{DEN}& \textbf{PARA}& \textbf{EXP} \\
         \midrule
         & 0.2718 & 0.0541 & 0.1529 & 0.1144 & 0.1259 \\
         \bottomrule
    \end{tabular}
    \caption{\label{tab:bertscoreplot}Pearson Correlation coefficient ($\rho$<0.05) between BERTscore($a$, $\hat{a}$) and \texttt{log}($\textbf{C}_{df}(q)$) for OLMo-1b.}
    
\end{table}

\subsection{CLS Embbedings and Co-occurrence Probability}

We investigated the link between words embeddings and frequency in the pre-training corpus by conducting the following experiments:
\begin{itemize}
    \item Firstly, we computed the \textbf{cosine similarity score} between the embeddings of the simple or multi-word medical term to be explained or defined, and the embeddings of the answer generated by the language models. We used sentence transformer \texttt{paraphrase-MiniLM-L6-v2}. In order to do an exhaustive comparison, we dissolve the reference answer ($a$) and generated answer ($\hat{a}$) into all possible n-grams. Then, we compute the pairwise cosine similarity between 2, 3, 4 and 5 possible n-gram pairs.

    \item Secondly, we calculated the \textbf{probability score} between two document-frequency counts: 1) the frequency of the query term ($q$) in the corpora ($C$), and 2) the frequency of the term together with the generated answer of the language model in the corpus. We used the \texttt{WIMBD} tool to found the frequencies in terms of number of documents from DOLMA, and \texttt{infini-gram} library for The Pile.

\begin{equation}
    P_{cooccurence} = \frac{\textbf{C}_{df}(q|\hat{a}, C)}{\textbf{C}_{df}(q|C)}
\end{equation}
\end{itemize}

We calculated different n-grams combinations, of 2 to 5 n-gram length. We kept the top-3 semantically meaningful values.
We show the statistical correlation between these variables using graphical representation.

\subsection{Tracing Back Head and Tail Knowledge}

We listed the most popular concepts (\texttt{head}) as those appearing most frequently in the parametric memory, and the least popular knowledge concepts (\texttt{tail}) as having the smallest number of corresponding documents in the parametric memory.

\begin{table*}
    \centering
    \resizebox{\textwidth}{!}{
    \begin{tabular}{lccp{13cm}}
        \toprule
        \rowcolor{blue!15}
        \textbf{Term} & \textbf{Docfreq} &  \textbf{Examples}\\
        \midrule
        disease	&75724477& A disease is a medical condition that affects the body's structure or function\\
        
cancer&50918098	&  A disease that affects the cells \\
anxiety &35107524& Anxiety is a feeling of fear and uneasiness\\
\midrule
testosterone-inhibiting	&115	&The answer is testosterone-inhibiting \\
biological tissue damage	&503	&Give examples of biological tissue damage\\
lifelong neurological	consequences &29	&Explain the neurological consequences of the disease\\
         \bottomrule
    \end{tabular}}
    \caption{\textcolor{black}{Examples of LLM paraphrasing. Frequent terms are paraphrased, while rare terms yield outputs similar to or derived from pre-training data.}}
    \label{tab:lm-semantic-paraphrasing}
\end{table*}

We traced back 100 concepts (1.62\%), where 50 were head concepts (0.8\%) and 50 percent tail concepts, to the pretraining corpus. We discarded the very long tail concepts from RefoMed-EN that had zero frequency. However, it is important to note that WIMBD's search is exact match based (while ignoring special characters and punctuation), and that some zero frequency terms are very long and technical, such as "disorder of bronchial ventilation" or "SMN1 gene-related proximal spinal muscular atrophy". To compare expert (EX) to general public (GP) datasets, we split this number evenly between RefoMed-EN-EX and RefoMed-EN-GP. We analyzed word level frequencies in 100 documents for each term downloaded with the WIMBD tool. 

As the linguistic diversity and the quality of the pretraining data is extremely important in the task performance evaluation, we conducted a close up analysis of corresponding documents in the DOLMA corpus. We analyzed the texts for a selected number of \texttt{head} and \texttt{tail} concepts.

\section{Results and Analysis}

\paragraph{Statistical correlations between medical concepts frequency in the parametric memory.  } 
\textcolor{black}{To answer \textbf{(RQ1)}, we computed the correlation coefficient between BERTscore and query term frequency \({C}_{df}\).}
We show the Pearson correlations scores coefficient between the BERTscore ($a$, $\hat{a}$) and log(\({C}_{df}\) (q)) on the full dataset of 6170 terms and gold paraphrases, in Table \ref{tab:bertscoreplot}. The best scores were obtained with OLMo-1b on the DEF query type (0.27), showing there is a slightly moderate correlation between the semantic similarity of the generated answer compared to the gold standard answer, and the frequency of the term in DOLMA. Second best scores are on DEN (0.15), while PARA and EXP have lower values (0.11 and 0.12). The lowest correlations score were obtained on the EX query type (0.05). \textcolor{black}{We obtained similar results for the bigger model, OLMo-7b, for DEF (0.23), DEN (0.15), and EXP (0.14). OLMo-7b has better scores for EX (0.16) and PARA (0.14), which shows that the model handles the complex linguistic task better. We conducted a manual analysis of the quality of the generated answers for each query type, described in the next section that tackles \textbf{(RQ2)}.}

When computed separately by type of corpus, expert medical texts (RefoMed-EN-EX) and general public medical texts (RefoMed-EN-GP), the results are consistent: the scores are the highest for DEF and DEN. However, the scores on RefoMed-EN-EX are better than those on RefoMed-EN-GP for DEF, and even higher than the values on the full dataset (0.31 compared to 0.24 and 0.27). This result indicates that in the expert texts, the task performance of the model on DEF is more correlated to the term frequency than in the general public (as RefoMed-EN-EX has a higher number of technical terms). On the contrary, for DEN, scores are higher on RefoMed-EN-GP (0.19) than RefoMed-EN-EX (0.11), also surpassing the full dataset score (0.15). We analyzed these results manually in the next section.

\paragraph{QA task performance evaluation on linguistically diverse query types.} 
To investigate \textbf{(RQ2)}, we conducted a qualitative analysis of the LLM's linguistic \textit{understanding}.
Our hypothesis was that language models show very high performance for the \texttt{definition} type query, as this type of query is very frequent in knowledge intensive QA benchmarks (\citeauthor{rebboud2024benchmarking}, \citeyear{rebboud2024benchmarking}; \citeauthor{zhao2024knowledgefmath}, \citeyear{zhao2024knowledgefmath}; \citeauthor{fei2023lawbench}, \citeyear{fei2023lawbench}). The results are consistent with our hypothesis (Table \ref{tab:bertscoreplot}). On the flip side, we observed that the LLM does not completely \textit{understand} the concept of \texttt{paraphrase} or \texttt{explanation}, as it generated definitions as answers instead.

We noticed that OLMo-1b obtained lower task performance with the \textit{denomination} and \textit{exemplification} types of queries, as it tends to either repeat the term (for denomination in particular) or give a definition style answer for both. The difficulties for these two queries come from linguistic and domain specific linguistic traits, such as:

\texttt{Denomination query} - the LLM repeats the term instead of giving another synonym for the medical concept. Furthermore, the presence of highly technical long-tail terms (such as \textit{lymphocytic bacterial meningitis}; freq=2) or opaque abbreviations (e.g.\textit{ FixM/F}; freq=11) renders the task even more difficult for the language model.

\texttt{Exemplification query} - the LLM does not output examples (i.e instances, types, subtypes) when the query term is very technical, thus long-tail knowledge: (\textit{Give examples of severe motor disorders} $\rightarrow$ \textit{The answer is severe motor disorders}). However, the model task performance increases for head knowledge (\textit{1. Give examples of cardiovascular disease} $\rightarrow$	\textit{a heart attack}; \textit{2. Give examples of psychological factors} $\rightarrow$ \textit{The answer is: Psychological factors include: fear, anxiety, and depression}).

\paragraph{Tracing back head and tail knowledge in the pretraining corpus.} 
We looked into a total of 5074 documents from the DOLMA corpus, where 5000 contained the head words and 74 the tail for the tail concepts. For each head term out of the 50, we downloaded 100 documents where this term appeared at least once. For tail terms, we looked for concepts that appeared in 1 to 3 documents in the DOLMA pretraining corpus.

The long tail terms with very low frequency (>4) are very technical long multi-word terms, such as "pseudo-rhizomelic polyarthritis", "lymphocytic bacterial meningitis", "CoA HMG reductase inhibitors". However, there are also very long multi-word terms in RefoMed-EN-GP, like "work-related musculoskeletal disorders of the upper limbs and neck" or "shiftworker sleep disorder" with a term frequency of 2 and 1 in the whole DOLMA corpus.

Regarding the head terms, the most frequent term from RefoMed-EN is "life", which appears in 497M documents in the pretraining corpus, while the second most frequent is "control" with 230M documents. Other head terms are "conditions" (130M), "function" (114M), "skills" (126M) and typical medical head concepts such as "treatment" (113M), "pain" (80M), "disease" (75M) and "cancer" (50M). While analyzing the term frequency, we noticed that terms such as "control", "function" and "client" appear in code texts, thus irrelevant for your medical knowledge QA task.

\begin{table}
    \centering
    \resizebox{\columnwidth}{!}{
    \begin{tabular}{c c c c c c c c}
        \toprule
        \rowcolor{blue!15}
        Criteria & \textbf{Model} &\textbf{EX}  && \textbf{GP} && \textbf{Refomed} \\
        \midrule
        Diversity &OLMo-1b& -0.3904  && -0.5710 && -0.4403\\
        &Pythia-1b & \textcolor{red}{-0.0206} && -0.1202&& -0.0317  \\
        Scalability & OLMo-7b & -0.3656&&	-0.3308&&	-0.3501 \\
        \bottomrule
    \end{tabular}}
    \caption{\textcolor{black}{Pearson correlation between CLS cosine similarity (query term, response-top3-ngrams) and the cooccurence probability of frequency of the term together with its top3 n-grams. (\textcolor{red}{Red} color denotes $\rho$>0.05)}  
    }
    \label{tab:pearson-CLS-coprob}
\end{table}

\paragraph{CLS n-gram embeddings on head and tail terms.}
\textcolor{black}{We explored \textbf{(RQ3)} in line with \citeauthor{wang2024generalization}'s (\citeyear{wang2024generalization}) hypothesis: LLMs generalize more in reasoning-intensive tasks, and they memorize more in simpler knowledge-intensive task (like in the case of tail knowledge).} We conducted the CLS n-gram analysis on the list of 100 head and tail concepts to verify this hypothesis. We calculated the CLS embedding BERTscore between the medical term and n-grams of the generated answer of different sizes (2, 3, 4 and 5 n-grams). We kept the top 3 best BERTscore for each term and we compared them with the frequencies of the term together with theses n-grams in the pretraining corpus. \textcolor{black}{We show the distribution of the results in Table \ref{tab:pearson-CLS-coprob}.}

The negative distribution scores (-0.44 on RefoMed-EN for OLMo-1b) indicate that the language model tends to create its own sentences and does not take full information package directly from DOLMA. \textcolor{black}{In terms of scalability, OLMo-7b is inline with OLMo-1b (-0.35 on RefoMed-EN), while another family of models, Pythia-1b, show very low distribution scores. This suggests that the LLM prioritizes semantic paraphrasing (backed by the semantic evidence in relation to the CLS similarity) (see examples in \Cref{tab:lm-semantic-paraphrasing}), and it is not reproducing the same content from the pretraining corpus, as proven by the syntactic evidence related to the n-gram occurrence. OLMo-1b's scores are better for head terms and for the general public texts (-0.71), showing that the model paraphrases more on popular and frequent knowledge and less on tail and technical knowledge, especially in the expert texts (+0.07), while OLMo-7b has similar values for both types of texts.} This indicates that the LLM memorizes more on knowledge intensive questions, as shown in previous studies \cite{wang2024generalization}.
\textcolor{black}{\section{Discussion}
\paragraph{Scaling up small language models does not necessarily improve performance.} We replicated our experiments on a bigger model, OLMo-7b. We notice a similar trend for definition-type questions: both OLMo-1b and OLMo-7b exhibit positive Pearson correlation coefficient (\(p\)) between BERTscore and query term frequency of 0.16. Interestingly, we noticed that OLMo-7b shows a weaker trend compared to OLMo-1b model (\(p\)=0.27), this implies that with increase in the size of language models the ability to deal with tail knowledge does not necessarily increase. Regarding overall comparison of CLS n-grams scores on head and tails (see Table \ref{tab:pearson-CLS-coprob}), we further notice a similar trend between OLMo-1b and OLMo-7b models.}

\begin{figure}
    \centering
    \includegraphics[width=\linewidth]{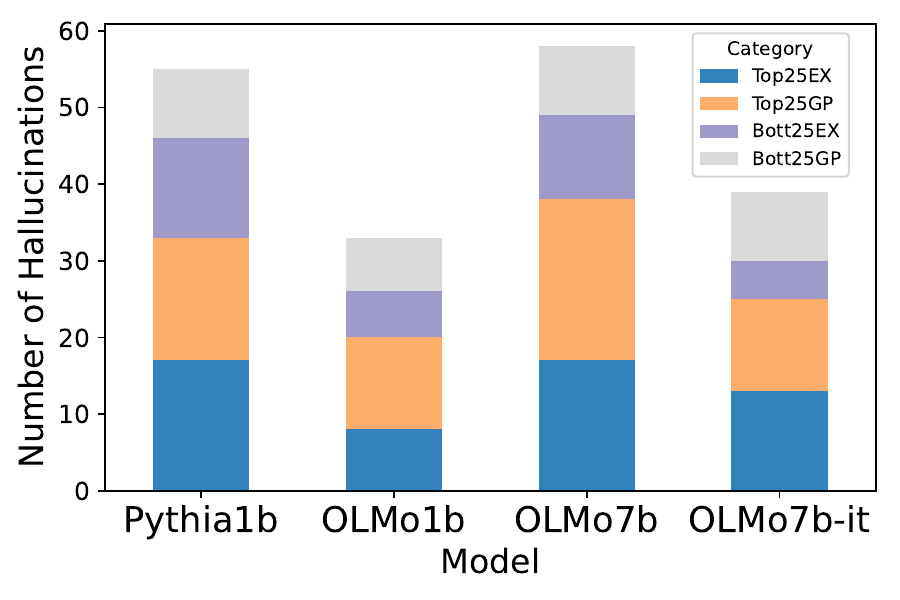}
    \caption{\textcolor{black}{Manual analysis of hallucinations on 100 head and tail terms. Best model is OLMo-1b, with the lowest rate of hallucinations (33\%).}}
    \label{fig:hallu_check}
\end{figure}

\textcolor{black}{\paragraph{Human evaluation shows hallucinations are not scale related.} We conducted a manual analysis of 400 answers given my the four models for the 100 head and tail terms dataset (Figure \ref{fig:hallu_check}). We evaluated task performance and diversity of models by looking directly into the data for hallucinations. Our analysis showed that Pythia-1b is more prone to generating hallucinatory texts (+22\%) compared to OLMo-1b, as it was previously shown \cite{groeneveld2024olmo}. As for bigger models, OLMo-7b is hallucinating more (+19\%) than its Instruct version. The best model remains OLMo-1b (33\% hallucinated answers), followed by OLMo-7b-Instruct (39\%). Pythia-1b demonstrates an opposite trend for definition-type questions (DEF) as compared to OLMo-1b, with a comparatively low Pearson correlation coefficient of -0.103. However, OLMo-7b shows a similar positive trend for paraphrase-type questions (PARA) with a coefficient of 0.14.}
\textcolor{black}{\paragraph{More linguistic diversity metrics are needed.} BERTscore might not be the best fit for all types of queries we analyzed. For example, for denomination-types queries where the LLM only repeats the medical term in the query, the BERTscore will be very high, but not relevant. In a concurrent work to our, \citeauthor{guo2024benchmarking} \citeyearpar{guo2024benchmarking} compared lexical, syntactic and semantic distribution of LLMs texts to human gold answers. \citeauthor{lee2023lftk} (\citeyear{lee2023lftk}) proposed a pipeline to identify 220 popular handcrafted linguistic features. However, the NLP community needs to continue working on the best linguistic features that count for each writing task.} 

\section{Conclusion}

\textcolor{black}{Our study introduced TrackList, a pipeline to analyze, trace back and evaluate a language model's answer to diverse linguistic queries. We showed that frequency of terms in the pretraining corpus impacts performance: LLMs tend to give inaccurate answers for head terms and more accurate answers for tail terms. LLMs are prone to paraphrase more on head, and thus popular, knowledge, and less on tail and technical knowledge, especially in the expert texts. Paraphrasing too much sometimes leads to inaccurate answers for head terms (like \textit{life, condition, disease}). Our study showed that Pythia-1b hallucinates more (+22\%) than the models from the OLMo family.
Our linguistic analysis showed that language models tend to give definition-type answers to different queries, even when asked to give examples or paraphrases. This demonstrates the limited linguistic knowledge of a small language model (1b-7b), still far from human knowledge.}

\section*{Ethical Considerations} The dataset used for this experiments has open license (CC BY-NC 4.0) and can be used for research by the NLP community. The RefoMed-EN dataset contains no personal data or patient data.

\section*{Limitations}

This work was conducted only on English using only fully open language models, OLMo \cite{groeneveld2024olmo} and Pythia \cite{biderman2023pythia}. Due to computational power limitations, we conducted our analysis on small language models (1b and 7b). Future studies could include the recently released OLMo2 family of models \cite{olmo20242} and testing new data tracing tools such as OLMoTrace \cite{liu2025olmotrace}. Other open source LLMs give access to their pretraining data, like BLOOM \cite{workshop2022bloom} available, but the big size of the dataset makes it difficult to explore.

Our study focused on small size language models to investigate the inner working of LLMs without domain knowledge finetuning, DPO finetuning or RAG systems. Our purpose was to analyze how the pretraining corpus and the word frequency (head and tail) impacts the accuracy of the LLMs answer to different linguistic types of questions. This is motivated by the fact that humans use language models as plug-and-play tools, without any finetuning methods. We are aware that using the methods listed above will improve vanilla LLM's performance for the QA task. Further research can include exploring these methods.
Moreover, we are aware of this limitation for your semantic similarity evaluation, and we further investigate medical  metrics such as MEDCON \cite{yim2023aci}, or fact checking metrics and tools such as FACTSCORE \cite{min2023factscore}, FIRE \cite{xie2024fire}, or LOKI \cite{li2025loki}. LLM-as-a-Judge evaluation method can also be explored and compared with existing tools and metrics.

\nocite{*}
\section{Bibliographical References}\label{sec:reference}

\bibliographystyle{lrec2026-natbib}

\end{document}